\documentclass[letterpaper]{article} 
\usepackage[draft]{aaai25}  
\usepackage{times}  
\usepackage{helvet}  
\usepackage{courier}  
\usepackage[hyphens]{url}  
\usepackage{graphicx} 
\usepackage{natbib}  
\usepackage{caption} 
\usepackage{algorithm}
\usepackage{algorithmic}
\usepackage{amsmath}
\usepackage{booktabs}
\usepackage{subfigure}
\usepackage{amsfonts}
\usepackage{newfloat}
\usepackage{cleveref}
\usepackage{listings}
\usepackage{tikz}
\usepackage{enumitem}
\usepackage{multirow}
\Crefname{equation}{Eq.}{Eqs.}
\Crefname{figure}{Fig.}{Figs.}
\Crefname{tabular}{Tab.}{Tabs.}
\Crefname{table}{Tab.}{Tabs.}

\DeclareCaptionStyle{ruled}{labelfont=normalfont,labelsep=colon,strut=off} 
\floatstyle{ruled}
\newfloat{listing}{tb}{lst}{}
\floatname{listing}{Listing}
\nocopyright

\newcommand{\overbar}[1]{\mkern 1.5mu\overline{\mkern-1.5mu#1\mkern-1.5mu}\mkern 1.5mu}

\newcommand{\method}[0]{MOPED}

\title{Multi-modal Pose Diffuser: A Multimodal Generative Conditional Pose Prior}
\author{
    Calvin-Khang Ta\textsuperscript{\rm 1}, Arindam Dutta\textsuperscript{\rm 1}, Rohit Kundu\textsuperscript{\rm 1}, Rohit Lal\textsuperscript{\rm 1}, Hannah Dela Cruz\textsuperscript{\rm 1}, Dripta S. Raychaudhuri\textsuperscript{\rm 1}, Amit Roy-Chowdhury\textsuperscript{\rm 2}
}
\affiliations{
    \textsuperscript{\rm 1}University of California, Riverside \\
    \{cta003,adutt020,rkund006,rlal011,hdela004,drayc001\}@ucr.edu \\
    \textsuperscript{\rm 2}University of California, Riverside Department of Electrical and Computer Engineering\\
    amitrc@ece.ucr.edu
}

\begin{document}

\maketitle
\begin{abstract}

The Skinned Multi-Person Linear (SMPL) model plays a crucial role in 3D human pose estimation, providing a streamlined yet effective representation of the human body. However, ensuring the validity of SMPL configurations during tasks such as human mesh regression remains a significant challenge , highlighting the necessity for a robust human pose prior capable of discerning realistic human poses. To address this, we introduce \method: \underline{M}ulti-m\underline{O}dal \underline{P}os\underline{E} \underline{D}iffuser. \method{} is the first method to leverage a novel multi-modal conditional diffusion model as a prior for SMPL pose parameters.
Our method offers powerful unconditional pose generation with the ability to condition on multi-modal inputs such as images and text. This capability enhances the applicability of our approach by incorporating additional context often overlooked in traditional pose priors. Extensive experiments across three distinct tasks—pose estimation, pose denoising, and pose completion—demonstrate that our multi-modal diffusion model-based prior significantly outperforms existing methods. These results indicate that our model captures a broader spectrum of plausible human poses.

\end{abstract}


\section{Introduction}
\label{sec:intro}


Accurate human pose estimation is crucial for several downstream applications, including but not limited to, biometrics~\cite{zhu2023gait,dutta2024poise}, augmented reality applications~\cite{Marchand2016PoseEF}, and pose guided image synthesis~\cite{ma2017pose}. Despite its importance, achieving precise monocular 3D human pose estimation is challenging. This difficulty arises from the lack of depth information in a single image and the limited generalizability of current algorithms to real-world scenarios. Although state-of-the-art algorithms for monocular 3D human pose estimation \cite{hmrKanazawa17,li2022cliff,pymaf2021} show impressive results on several datasets \cite{h36m_pami,lin2014microsoft,andriluka14cvpr,mono-3dhp2017}, they struggle with real-world conditions such as occlusions, varying illumination, and motion blur \cite{zhang2019unsupervised}. While deep learning models are known to benefit from data scaling \cite{rombach2021high}, this is particularly challenging for 3D pose estimation due to the difficulty and expense of capturing large-scale ground truth data in practical, real-world settings. 
This limitation hampers the generalizability of existing algorithms.

 \begin{figure}[t]
     \centering
     \includegraphics[width=.9\linewidth]{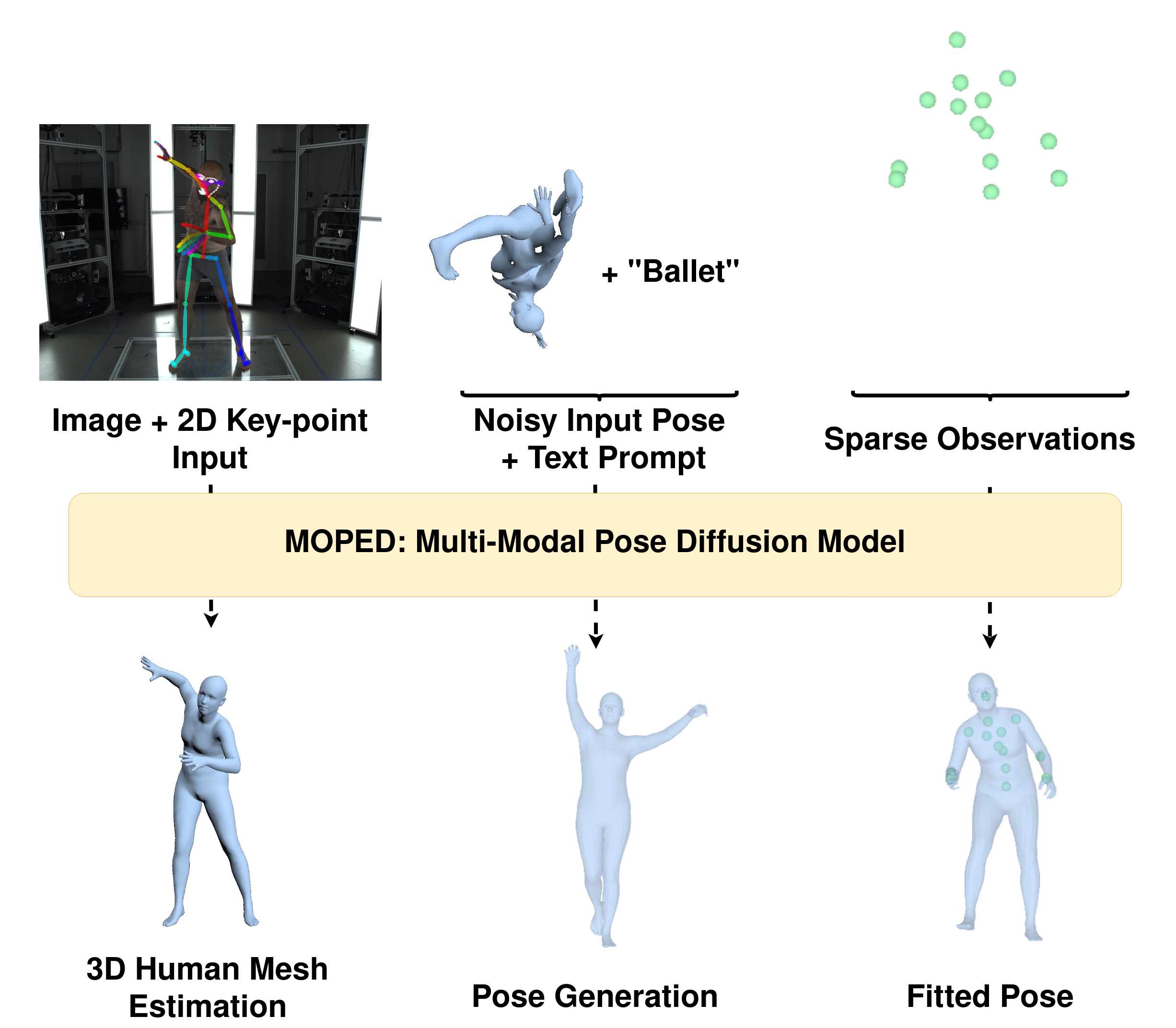}
     \caption{\method{} is a flexible model that has learned to accurately represent realistic poses and condition itself on multiple modalities. It is capable of being deployed across a variety of tasks. In this work we specifically showcase results across 3D human mesh estimation, pose generation, and pose completion. 
     }
     \label{fig:teaser}
 \end{figure}

The absence of labeled 3D data can potentially be offset by integrating the inherent bio-mechanical constraints that humans naturally follow. Human pose priors, which encapsulate these bio-mechanical constraints, are essential for ensuring the plausibility and validity of poses generated by pose estimation algorithms. Earlier works \cite{parameswaran2004view,bogo2016keep} imposed limitations on limb lengths and joint extensions to maintain realistic poses. However, these priors are often too simplistic to model the diversity of human poses. With the advent of large motion capture datasets \cite{AMASS:ICCV:2019}, attention has shifted towards data-driven priors. Past research has attempted to model the distribution of poses using Gaussian Mixture Models \cite{bogo2016keep}, Variational Auto-Encoders \cite{SMPL-X:2019}, Generative Adversarial Networks \cite{davydov2022adversarial}, and  Neural Distance Fields \cite{tiwari2022pose}. Although these pose priors have shown promise in improving the accuracy of pose estimation tasks, they often operate with limited contextual information. Given the ill-posed nature of 3D pose estimation, these priors, while focusing on the plausibility of the pose estimates, lack the ability to incorporate semantic information such as images and/or text. This often results in poses that appear accurate from a single view but are inaccurate from other perspectives.  
\emph{We propose that by incorporating additional contextual information from images and/or text, we can condition pose priors in a way that reduces the sample space to more semantically relevant poses.}

Inspired by the robust performance of diffusion models in generation tasks and their capability to easily integrate external conditioning information, we introduce the first multi-modal pose diffusion model, named \underline{M}ultim\underline{O}dal \underline{P}os\underline{E} \underline{D}iffuser (\method). \method~\emph{is a novel diffusion-based approach designed to model the distribution of plausible human poses while incorporating contextual information from various modalities such as images and language.} \method~enables multi-modal conditioning, providing maximum flexibility for deployment across diverse settings. It excels in accurately generating human poses in both unconditional and conditional scenarios involving text or images, demonstrating its potential as a versatile generative tool in human pose modeling. Fig. \ref{fig:teaser} illustrates the versatility of \method~through various downstream applications. In Fig. \ref{fig:hmr_examples}, we show some preliminary results on how improved pose priors can lead to more accurate 3D pose estimation.


\noindent{\textbf{Contributions:}} We make the following contributions in this work. 
\begin{itemize}[topsep=0pt]
\item  We introduce \method, a versatile generative human pose prior that models the distribution of natural and realistic human poses. 
\item  \method~is the first pose prior that features a flexible conditioning mechanism that integrates information from images and language, enabling the learning of semantically accurate human poses. 
\item  \method~demonstrates strong performance across various pose-oriented tasks, including 3D human mesh estimation, pose generation, and pose completion.
\end{itemize}
\begin{figure}[t]
    \centering
        \includegraphics[width=\linewidth]{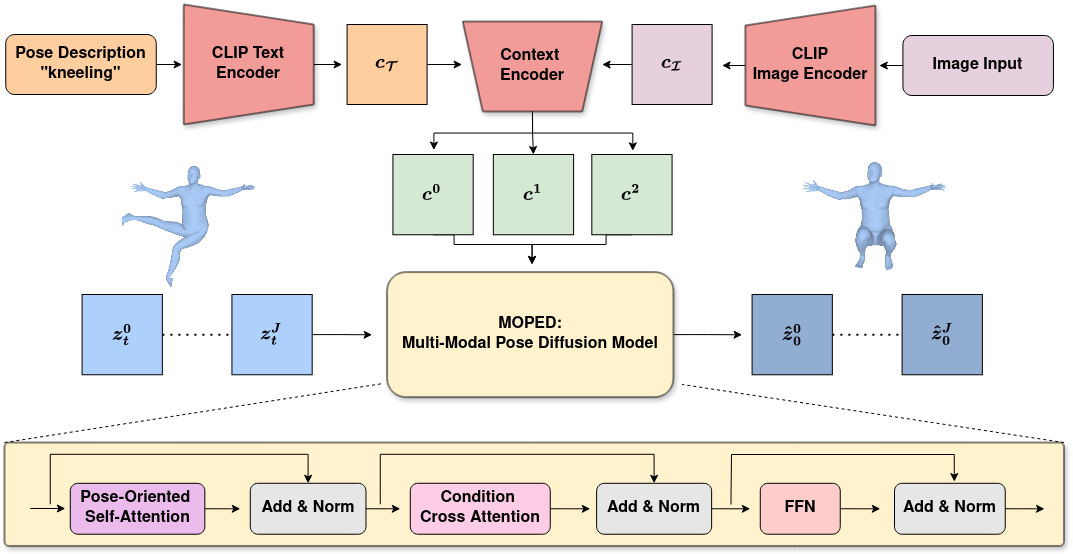}
        \caption{The architecture of \method{} is a transformer based model designed to specifically model the intra-joint relationship through the Pose-Oriented Self-Attention \cite{li2023pose}. Given some an input pose, $z_t$, we pass that through the our model $\mathcal{G}$, which consists of blocks containing a Pose-Self Attention layer \cite{li2023pose} and a cross-attention \cite{vaswani2017attention} mechanism which allows the context, $c$, to be incorporated into the model. This gives us the estimated pose, $\hat{z}_0$. \method{} is flexible to the input and can be conditioned on multiple modalities, (images and/or natural language), or sampled unconditionally.}
        \label{fig:diffusion_arch}

\end{figure}


\section{Related Works}
\label{sec:relwork}

\noindent{\textbf{Monocular 3D Human Pose Estimation:}} The task of monocular 3D human pose estimation is a fundamental problem in the field of computer vision which involves locating the keypoints such as head, shoulders and other landmarks on the human body in the three dimensional coordinate system from a single image. State of the art algorithms primarily fall into either regression based approaches or optimization based approaches. Regression based approaches rely on either directly estimating the 3D coordinates of the joints on the human body in a single image as discussed in works like \cite{martinez2017simple, gong2023diffpose, 10204476, hmrKanazawa17, li2022cliff, goel2023humans, pymaf2021,lin2021end-to-end,Moon_2020_ECCV_I2L-MeshNet}. Regardless, all of these algorithms usually rely on large scale supervised training on large amounts of paired data and often find it hard to generalize to real-world situations. On the other hand, optimization based approaches like SMPLify~\cite{bogo2016keep, SMPL-X:2019} solve for the 3D coordinates conditioned on the 2D keypoints of the human in the image. However, these methods suffer from high latency during inference but are more robust to out-of-distribution scenarios and lend themselves to be a natural place to include a prior~\cite{tiwari2022pose, SMPL-X:2019, davydov2022adversarial, lu2023dposer}. \\

\noindent{\textbf{Human Pose Priors:}} Monocular supervised human pose estimation methods often predict implausible poses when dealing with depth-ambiguity or out of domain scenarios. Pose priors can play a critical role in constraining the solution space and improving the robustness of pose estimation algorithms~\cite{bogo2016keep,SMPL-X:2019,raychaudhuri2023prior, stathopoulos2024score}. Pose priors have been extensively studied~\cite{parameswaran2004view,felzenszwalb2005pictorial, andriluka2009pictorial} in the literature, ranging from simple anatomical constraints to more sophisticated graphical models. However with the rise of deep learning and readily available datasets, learning priors from data has been a popular method for increasing their accuracy and performance. Previous works in the space have covered a wide range of approaches such as Gaussian Mixture Models (GMMs)~\cite{bogo2016keep}, Variational Auto Encoders(VAE)~\cite{SMPL-X:2019}, Generative Adversarial Networks (GANs)~\cite{davydov2022adversarial}, and Neural Distance Fields~\cite{tiwari2022pose, he2024nrdf}.

In this work we aim to explore modeling the distribution of poses using diffusion models. Diffusion models \cite{sohl2015deep,ho2020denoising,song2021scorebased} are a class of generative models that have gained significant attention in the field of machine learning for their ability to generate high-quality, diverse data samples. These models operate by simulating a diffusion process, which involves gradually adding noise to data and then learning to reverse this process to generate new data samples. Previous diffusion based works such as GFPose~\cite{10204476} and DPoser~\cite{lu2023dposer} aim to model the joint coordinates and SMPL parameters respectively. While these models have demonstrated promising results, they do not incorporate additional conditioning information beyond observed poses. Closest to our work is ScoreHMR~\cite{stathopoulos2024score} which features a diffusion based approach for refining human poses conditioned on image features. While ScoreHMR offers great performance, it requires an initialization from a prior pose estimation model, as well as image features, and as such is not usable for other tasks outside of image-based pose tasks. In contrast, we aim to leverage additional image and/or text information as \emph{optional} conditioning information allowing for a much more flexible pose prior for tasks beyond image based pose estimation such as pose generation or pose completion.  

\section{Methodology}
\label{sec:method}

In this section, we introduce our proposed algorithm \method{}, a conditional generative model ($\mathcal{G}$) that learns the distribution of realistic human poses. \method{} is a versatile generative model which can accommodate different types of conditioning inputs, such as images ($\mathcal{I}$) and/or natural language ($\mathcal{T}$) to produce semantically accurate human poses ($\theta \in \mathcal{R}^{24\times 6}$). Mathematically, the pose is computed as $\theta = \mathcal{G}(\mathcal{Z}; \mathcal{I}, \mathcal{T})$ where $\mathcal{Z} \in \mathcal{R}^{24\times6}$ which can take the form of either pure Gaussian noise or an estimated pose from any source. Note that, different from existing human pose priors \cite{SMPL-X:2019,tiwari2022pose,lu2023dposer}, \method{} \emph{leverages both textual and image features while learning the distribution of human poses}. Without loss of generality, we utilize the SMPL \cite{10.1145/2816795.2818013} model for learning \method{} ($\mathcal{G})$. Fig. \ref{fig:diffusion_arch} provides an overview of our proposed algorithm \method. \\

\noindent {\textbf{SMPL Model:}} The SMPL \cite{10.1145/2816795.2818013} model provides a differentiable function that, when given a pose keypoints ($\theta \in \mathcal{R}^{24\times6}$) and shape parameters ($\beta \in \mathcal{R}^{10}$), computes a mesh $V\in\mathcal{R}^{6890\times3}$ . The pose parameters ($\theta \in \mathcal{R}^{24\times6}$) consists of a global orientation of the pelvis (root joint) with respect to an arbitrary coordinate system and 23 relative joint rotations with respect to a parent joint along the skeleton. Specifically, we utilize a continuous 6D representation \cite{Zhou_2019_CVPR} for representing the joint rotations, which has been shown to result in better performance for end-to-end models when dealing with rotations. 

\label{subsec:model_arch}

\subsection{Architecture of \method}
\label{subsec:model_arch}

We propose to use a diffusion model \cite{ho2020denoising, song2021scorebased, rombach2021high} as the backbone of \method, to model the distribution of realistic human poses. The architecture of \method{} is broadly split into two major components: the diffusion model $\mathcal{G}$ and the condition fusion module $\mathcal{F}$. In this section, we elaborate upon the architectural design of both the diffusion model $\mathcal{G}$ and the condition fusion module $\mathcal{F}$, with Fig. \ref{fig:diffusion_arch} showing the overall architecture of \method. \\

\noindent\textbf{Diffusion Model}
 The diffusion model ($\mathcal{G}$) utilizes a transformer-based architecture \cite{vaswani2017attention}, drawing significant inspiration from the Pose-Oriented Transformer \cite{li2023pose}. The inputs to $\mathcal{G}$ are linear projections of the input, $\mathrm{x}_t = \texttt{Proj}(z_t) \in \mathcal{R}^{J\times D}$, where $J$ is the number of joints and $D$ is the size of the latent dimension. Each element in $\mathrm{x}_t$ incorporates learnable positional embeddings $K$ which represents the absolute position of each joint. Additionally, we use a group position embedding $G\in\mathcal{R}^{5 \times D}$~\cite{li2023pose}, where each element of the input is divided into 5 groups based on their distance from the pelvis. The group embedding is used to represent groups of joints and allows for improved modeling between joints that lie further away from the pelvis. The embedding used is determined by the shortest skeletal distance of a group from the pelvis. The final input for our transformer model is defined as $\mathrm{x}_t^i = \mathrm{x}_t^i + K^i + G^{\phi(i)}$, where $i$ denotes the index of the input joint and $\phi(i)$ represents the distance of joint $i$ from the root joint. For simplicity, we will omit the time step notation when referring to the model input going forward.

Alongside the pose-oriented embeddings, we apply Pose-Oriented Self-Attention \cite{li2023pose}, a variant of the self-attention mechanism \cite{vaswani2017attention} that explicitly models intra-joint relationships. The attention matrix $A$ enables joints to weight the features of other joints based on the skeleton topology and is defined as follows:
\begin{align}
    A_{i,j} = \frac{(\mathrm{x}^iQ)(\mathrm{x}^jK)^T}{\sqrt{d}} + \Phi(\mathcal{D}(i,j)),
\end{align}
where $Q$ is the query projection matrix, $K$ is the key projection matrix, $\mathrm{x}$ represents the projected joints, $\mathcal{D}$ is the distance between joints $i$ and $j$ on the skeleton, and $\Phi$ is a network that projects the distance into the dimension $H$, which is the number of heads in the attention mechanism. Pose-Oriented Self-Attention \cite{li2023pose} helps our model adjust the contribution of other joints based on their distance, rather than treating all joints equally. \\

\noindent\textbf{Conditioning Model:}
Unlike previous works \cite{tiwari2022pose, he2024nrdf, lu2023dposer}, our approach aims to enhance \method{} by incorporating additional context through various modalities, specifically natural language \textbf{and/or} images. By integrating this additional context, we intend to condition the prior on critical information that cannot be inferred from the estimated pose alone. To incorporate images or natural language as conditioning data, we utilize a frozen CLIP ~\cite{radford2021learning} model which provides a rich shared semantic latent space. 

We denote the CLIP text encoder as $\mathcal{E}_\mathcal{T}$ and the CLIP image encoder as $\mathcal{E}_\mathcal{I}$, with their outputs represented as $c_\mathcal{T}\in \mathcal{R}^{D}$ and $c_\mathcal{I} \in \mathcal{R}^{D}$. To fuse $c_\mathcal{T}$ and $c_\mathcal{I}$, we use a vanilla transformer encoder to generate our conditioning tokens $c = \mathcal{F}(\mathcal{C},\mathcal{E}(\mathcal{I}), \mathcal{E}(\mathcal{T}))$
where $\mathcal{C}$ is a learnable \texttt{CLS} token which effectively aggregates information from the rest of the inputs. We than need to embed our time step $t$ by projecting the time step $t$ into a $D$-dimensional vector, then sum the time step $t$ and the condition $c$. We integrate $c$ into the diffusion model using cross-attention \cite{vaswani2017attention} between $\mathrm{x}$ and $c$, and decode it into the sampled pose $\hat{z}$.





\subsection{Training}
\label{subsec:difftrain}
Following standard training procedures for training Denoising Diffusion Probablistic Models \cite{ho2020denoising} (DDPM), the forward process is represented as a Markov Chain,  where noise is gradually added to the original pose data $z_0$ over $T$ transitions (we use $T=1000$). At any timestep $t$, the forward process is  expressed as, $$q(z_{t} | z_0) = \mathcal{N}(z_t;\sqrt{\overbar{\alpha}_t}z_0,(1-\overbar{\alpha}_t)\mathbf{I}),$$ where $\beta_t$ is fixed variance schedule with $\alpha_t = 1-\beta_t$ and $\overbar{\alpha}_t = \prod_{i=0}^t  \alpha_i $. At any timestep $t$, a noisy sample is $z_t = \sqrt{\overbar{\alpha}_t}z_0 +\sqrt{1-\overbar{\alpha}_t}\epsilon$ where $\epsilon \sim \mathcal{N}(0,\mathbf{I})$. The reverse process $q(z_{t-1} | z_t, z_0)$ is the process that gradually removes the noise $\epsilon_t$ from $z_t$ until we reach the original data distribution at $z_0$.

In this work, we aim to learn a model $\mathcal{G}(z_t; c)$ that can denoise some arbitrary noisy pose sample $z_t$ back to $z_0$ given some condition $c$. The objective function that optimizes our model is the simplified objective described in the original DDPM \cite{ho2020denoising} algorithm.
\begin{equation}
    \mathcal{L} = 
    \mathbb{E}_{z_0 \sim p_\text{data}, t\sim[1,T]} 
    [||\epsilon_t - \mathcal{G}(z_t;c)||^2_2]
    \label{eq:1}
\end{equation}

With Eq. \ref{eq:1} as the loss function, we train $\mathcal{G}$ on a large distribution of realistic human poses, which are denoted by $p_\text{data}$\cite{AMASS:ICCV:2019,BABEL:CVPR:2021}. Moreover, \method{} undergoes training with classifier-free guidance\cite{ho2021classifier}, where the conditioning input $c$ is randomly omitted $10\%$ of the time, enabling the sampling of $\mathcal{G}$ without consideration of the available modalities.


\subsection{Sampling}
In order to sample from the diffusion model we employ denoising diffusion implicit model (DDIM) \cite{song2020denoising} based sampling for all of our experiments. DDIM allows for accelerated sampling by taking larger steps and defines sampling $x_{t-1}$ from $x_t$ as follows:
\begin{align}
    x_{t-1} = \sqrt{\overbar{\alpha}_{t-1}}\hat{x}_0 + \sqrt{1-\overbar{\alpha}_{t-1}-\sigma^2_t} \cdot \mathcal{G}(z_t;c) + \sigma_t\epsilon
    \label{eqn:ddim}
\end{align}
where $\hat{x}_0$ is the estimated starting sample which is defined as 
\begin{align}
        \hat{x}_0 = \frac{x_t - \sqrt{1-\overbar{\alpha_t}}\mathcal{G}(z_t;c)}{\sqrt{\overbar{\alpha}}_t}
        \label{eqn:est_start}
\end{align} Deterministic sampling is than enabled by setting $\sigma_t=0$. 

\section{Experiments}
\label{sec:exp}


In this section we describe the experimental setup and metrics used to evaluate \method{}. We examine \method{} under the tasks of human mesh regression, pose denoising, and pose completion.

\subsection{Implementation Details}
We first conduct a \textit{pretraining} phase in which we only sample poses from the AMASS training split \cite{AMASS:ICCV:2019} which is a large collection of motion capture data. Additionally we leverage the textual labels from the BABEL \cite{BABEL:CVPR:2021}. This allows us to leverage a large amount of human pose data without the need to train on images and allows our model to learn a distribution of plausible human poses. In the second phase of training, we fine-tune our model on a mix of standard 2D and 3D pose estimation datasets: Humans3.6M \cite{h36m_pami}, MSCOCO \cite{lin2014microsoft}, MPII \cite{andriluka14cvpr}, MPI-INF-3DHP \cite{mono-3dhp2017}. In order to source captions for these images, we either use the supplied captions, if available, or we generated the captions using off the shelf automatic image captioning\cite{li2022blip}. All of our models are trained on a single Nvidia RTX 3090 with 1 million iterations of pretraining and 10K iterations on joint image-text fine-tuning. 
\subsection{Generation}
\begin{figure*}[]
    \centering
    \includegraphics[width=.8\linewidth]{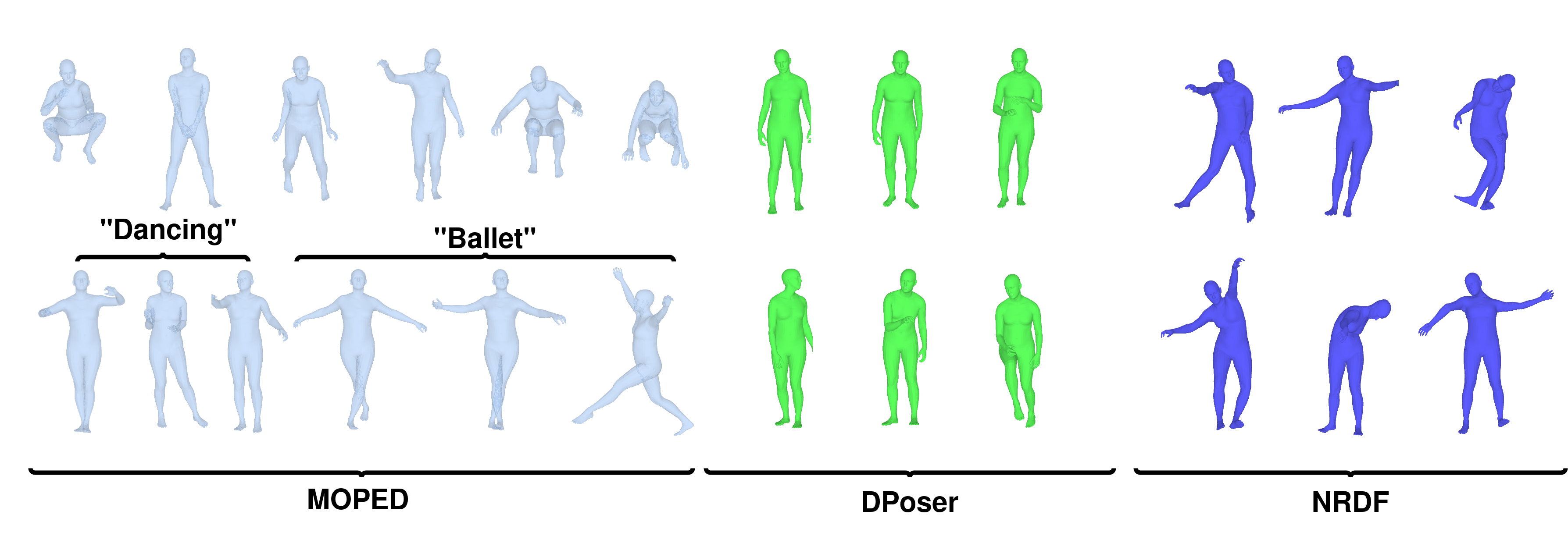}
    \caption{On the left hand side we see that \method{} is able to both unconditionally (top row) and conditionally (bottom row) generate realistic poses with significant diversity. In contrast we observe that DPoser tends to produce realistic poses with much more limited diversity. In the case of NRDF, we find that while it exhibits much greater diversity, it suffers from significantly less realism. Additional examples can be found in the Supplemental Material.}
    \label{fig:Generation Examples}
\end{figure*}

\begin{table}[b!]
\centering
\setlength{\tabcolsep}{1mm}
\begin{tabular}{lccc}
\toprule
\textbf{Method}                     & \textbf{FID} $\downarrow$ & \textbf{APD}$\uparrow $ &$ d_\text{NN} \downarrow $\\
\midrule
GMM \cite{bogo2016keep}         & 0.435 & 21.944 & 0.159\\
VPoser \cite{SMPL-X:2019}       & 0.048 & 14.684 & 0.074 \\
PoseNDF\cite{tiwari2022pose}    & 3.920 & 37.813 & 0.838 \\
DPoser\cite{lu2023dposer}       & 0.027 & 13.843 & 0.075 \\
NRDF\cite{he2024nrdf}           & 0.636 & 23.116 & 0.177 \\
\method{} (Ours)                & 0.200 & 20.559 & 0.145 \\
\bottomrule
\end{tabular}
\caption{We report the FID, APD, and $d_\text{NN}$ metrics for generation quality. When compared to previous works we are able to maintain improved realism while maintaining strong diversity.}
\label{Table: Generation}
\end{table}


\label{subsec:generation}
In order to evaluate the generative performance of \method{}, we employ DDIM based sampling with a step size of $50$. We look specifically at the realism and diversity of the samples generated and compare against previous works. In line with experiments in NRDF \cite{he2024nrdf} we also evaluate using the same metrics. The first is the Fréchet inception distance (FID) which compares the distribution between the generated samples and the training set. We also employ the $d_\text{NN}$ metric as proposed in NRDF~\cite{he2024nrdf} which is the quaternion geodesic distance to the nearest neighbor in the training dataset. 
Finally, we utilize the Average Pairwise Distance(APD)~\cite{aliakbarian2020stochastic} to evaluate the diversity of the generated poses. We compare \method{} against previous works capable of unconditional generation such as GMM \cite{bogo2016keep}, VPoser \cite{SMPL-X:2019}, PoseNDF \cite{tiwari2022pose}, DPoser \cite{lu2023dposer}, and NRDF \cite{he2024nrdf}. From Table \ref{Table: Generation}, we can see that we are able to exceed SOTA works such as NRDF in terms of FID and $d_\text{NN}$ indicating that we are able to generate more realistic poses. In addition \method{}, is able to generate diverse poses in terms of APD when compared to models with extremely low FID or $d_\text{NN}$ such as VPoser or DPoser which tend to generate mean poses. Visually in Fig. \ref{fig:Generation Examples} we can see that \method{} is ultimately able to effectively maintain a good balance between realism and diversity. However, \emph{what makes \method{} stand out from \textit{all} other priors is its ability to conditionally generate poses based off or text and/or images which can improve performance on downstream tasks.}
%
        

    \begin{table}
        \centering
        \setlength{\tabcolsep}{1mm}
        \begin{tabular}{lll}
        \toprule
         & EHF  &3DPW \\
         \cmidrule{2-3}
        \textbf{Method}   & PA-MPJPE & PA-MPJPE \\
        \midrule
        HMR2.0 & 51.09 & 54.3 \\
        \midrule
        VPoser\cite{SMPL-X:2019}   & 58.08  / 72.79  & 133.53/70.27  \\
        Pose-NDF\cite{tiwari2022pose} & 57.87 /76.33    & 137.16/74.08 \\ 
        NRDF \cite{he2024nrdf} & N/A   &  68.47* \\
        DPoser \cite{lu2023dposer} & 56.05 / 71.523   & 135.50/68.87 \\
        \method{} (Ours)   & 49.05    & \textbf{50.78} \\
        \midrule
        ScoreHMR & \textbf{49.03} & 51.1\\
        \bottomrule
        \end{tabular}        
        \caption{
        We compare \method{} with other pose priors on both the EHF and 3DPW datasets. For pose priors that utilize the SMPLify framework, we test both a default initialization and with HMR2.0 respectively. We separate out ScoreHMR as, while it is a diffusion based 3D pose refinement method, it requires an image as an input and is \textit{not} capable of other tasks we describe where the image is not available. *At the time of writing, NRDF does not have publicly available code for Human Mesh Regression and the reported results for 3DPW are on a \textbf{subset} of the 3DPW test split as described in their paper, while we report results for the entire test split.}
        \label{table:hmr}
    \end{table}

\begin{figure*}
    \centering
    \includegraphics[width=.7\linewidth]{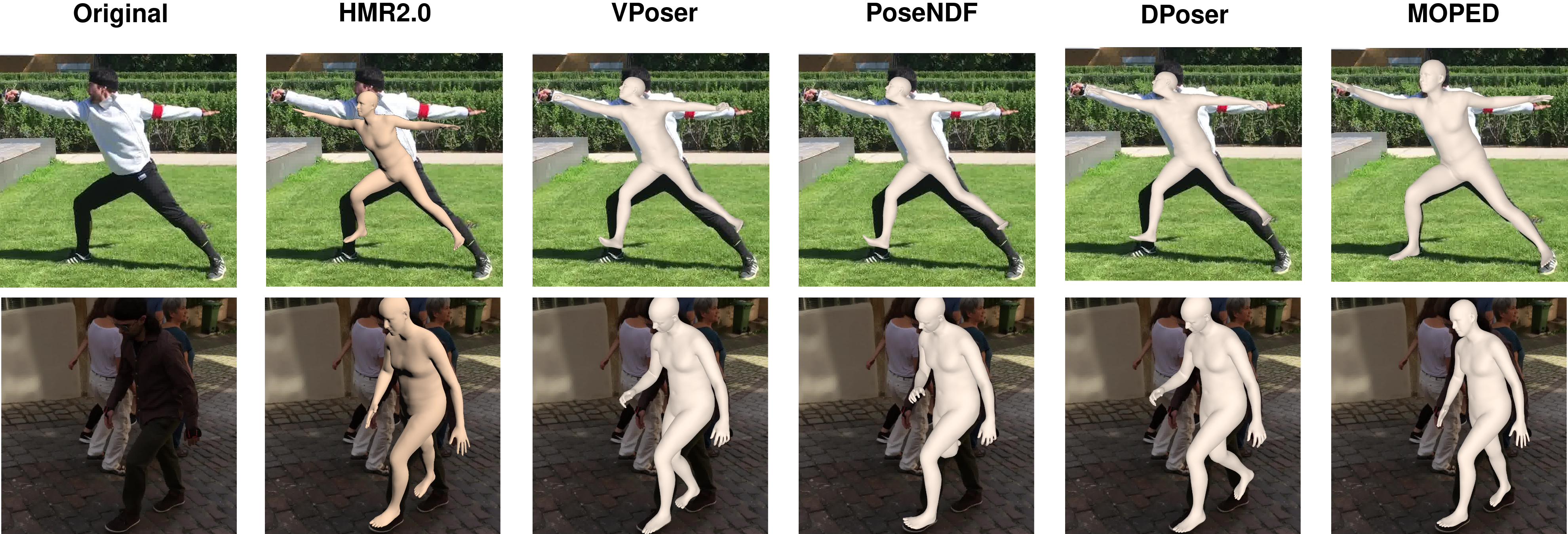}
    \caption{Compared to other pose priors, \method{}~is able to effectively leverage the HMR2.0 initialization. On the top row we see that \method{}~ results in far better fitting than the other pose priors. On the bottom row we see that \method{} results in improved alignment with the right leg while maintaining the right arm alignment.  Additional results with larger images are in the Supplemental Material.}
    \label{fig:hmr_examples}
\end{figure*}
\subsection{Human Mesh Recovery}
\label{subsec:applications_opt}
In order to demonstrate the effectiveness of \method{} as a prior, we use the task of Human Mesh Recovery. Specifically, we demonstrate \method{}'s ability to work in both unconditional and conditional settings and how the incorporation of both text and images results in improved performance. We show results on both the EHF \cite{SMPL-X:2019} and the 3DPW datasets \cite{vonMarcard2018} and compare against previous works. We evaluate performance using the standard Procrustes Aligned Mean Per Joint Position Error (PA-MPJPE) metric as done in previous works. 
The typical setup for Human Mesh Recovery using pose priors is based off of SMPLify~\cite{SMPL-X:2019,bogo2016keep}, an optimization based framework which aims to minimize the following objective: 
\begin{align}
    \mathcal{L}_\text{SMPLify} = \mathcal{L}_\text{Data} + \lambda_1\mathcal{L}_\text{Prior} + \lambda_2\mathcal{L}_\text{Shape} + \lambda_3\mathcal{L}_\text{Angle},
\end{align}
where the data term is a reprojection loss defined as 
\begin{align}
    \mathcal{L}_\text{Data} = \sum_\text{joint i}\gamma_i\omega_i\rho(\Pi_K(R_\theta(J(\beta))_i - J_\text{est,i}))
    \label{eqn:data}
\end{align}
where $J_\text{est}$ are estimated key-points provided by OpenPose \cite{8765346}, $R_\theta(J(\beta))$ is the 3D key-points from the SMPL model, $\Pi_K$ is a 3D to 2D projection with intrinsic camera parameters $K$, and $\rho$ is the Geman-McClure error \cite{geman1987statistical}.
The $\mathcal{L}_\text{Angle}=\exp{}(z_\text{est})$ term penalizes extreme angles for the elbow and knee joints and the $\mathcal{L}_\text{Shape}=||\beta||^2$ term penalizes large shape values. 

While \method~is effective in the SMPLify framework similar to previous works (additional results available in the supplemental material), there are limits to the improvement especially in the presence of a more accurate initialization. The optimization methods are much more sensitive to hyper-parameters and often can result in worse performance. In the presence of more accurate models such as HMR 2.0 \cite{goel2023humans}, the SMPLify approach tends to exacerbate errors as similarly observed in ScoreHMR~\cite{stathopoulos2024score}. Instead of using the SMPLify framework, we can instead sample the posterior $p(x_t | c, y)$~\cite{chung2022diffusion, stathopoulos2024score} which allows us to integrate feedback from loss functions directly into the sampling process. In order to correct the pose from an off-the-shelf pose estimation model and similar to works in image editing, we employ DDIM \cite{song2020denoising} inversion. DDIM inversion is the process in which we take some sample $x_0$ and invert the sampling process back to some timestep $t$. 

Following ScoreHMR~\cite{stathopoulos2024score}, we can use the loss term $\mathcal{L}_\text{Data}$ to guide the sampling process. Let $x_\text{init}$ be the initial estimate of the SMPL parameters from an image $\mathcal{I}$ provided by an off-the-shelf pose estimation method such as HMR2.0~\cite{goel2023humans}. In order to sample a new estimate, we first employ DDIM to invert $x_\text{init}$ to some chosen time step $t$. We are interested in sampling from $p(x_t | c, y)$ where $c$ is our conditioning information (image and/or text) and $y$ is our observation in the form of 2D key-points. In the diffusion model framework, the score of $p(x_t | c, y)$ is necessary for sampling and by using Bayes Rule the score can be decomposed as:
\begin{align}
    \nabla_{x_t}\text{log}(p(x_t | c, y)) &=  \nabla_{x_t}\text{log}(p(x_t | c) p(y |x_t, c))
\end{align}
where $\nabla_{x_t}\text{log}(p(x_t | c))$ is defined in terms of the noise $\epsilon$ estimated from $\mathcal{G(\cdot)}$.
The posterior $\nabla_{x_t} p(y | x_t, c)$ is intractable and thus must be approximated. Following ScoreHMR \cite{stathopoulos2024score} and Diffusion Posterior Sampling~\cite{chung2022diffusion} the posterior can be approximated as
\begin{align}
    \nabla_{x_t} \text{log} p(y| c, x_t) &\simeq    \nabla_{x_t} \text{log} p(y| c, \hat{x}_0)\\
    &= \nabla_{x_t}|| y - \mathcal{L}_{\text{Data}}(\hat{x}_0)||^2_2,
\end{align}
where $y$ are the observed 2D key-points and $\hat{x}_0$ is the estimated original sample defined in Eqn. \ref{eqn:est_start}. Using the estimated score conditioned on the 2D key-points this can be applied to DDIM sampling by adding the score to estimated noise $\epsilon_t$. This results in redefining $\mathcal{G}(z_t;c)$ as the following
\begin{align}
    \mathcal{G}_\text{DPS}(z_t;c) = \mathcal{G}(z_t;c) + \rho\sqrt{1-\overbar{\alpha}_t}\nabla_{x_t}|| y - \mathcal{L}_{\text{Data}}(\hat{x}_0)||^2_2
\end{align} where $\rho$ is a scaling hyper-parameter which we set to $\rho=0.003$.
This can than be plugged into the DDIM sampling step in Eqn. \ref{eqn:ddim}. In order to apply loss guided sampling to pose estimation, we first invert the output of HMR2.0 to $\hat{x}_t$ where $t=50$ then we denoise back to $t=0$ with a step size of $\Delta2$. 

We evaluate loss-guided sampling on both the EHF and 3DPW datasets in Table \ref{table:hmr} and show that \method{}~exceeds the performance of state-of-the-art pose priors. It is also able to offer competitive performance when compared to state-of-the-art pose refinement methods such as ScoreHMR which requires an image as an input and is \textit{not} capable of other tasks we describe where the image is not available.
Visually in Fig. \ref{fig:hmr_examples} we can observe that visually \method{} conditioned on images and text is able to refine the estimates provided by HMR2.0 much more accurately than other traditional pose priors. We include additional experiments regarding the step size and the choosing $t$ in the Supplementary Material.

\begin{table*}[!htb]
\begin{tabular}{l|ccc|ccc|ccc}
\hline
                                               &                                                        & Occ Arm                                              &                               &                                                        & Occ Legs                                             &                               &                                                        & End Effectors                                        &                               \\
\textbf{Method}                                & \textbf{FID}$\downarrow$ & \textbf{APD}$\uparrow$ & $d_\text{NN} \downarrow$ & \textbf{FID}$\downarrow$ & \textbf{APD}$\uparrow$ &$d_\text{NN} \downarrow$ & \textbf{FID}$\downarrow$ & \textbf{APD}$\uparrow$ & $d_\text{NN} \downarrow$\\ \hline
Pose-NDF~\cite{tiwari2022pose} & 1.460        & 16.445      & 0.622 &     3.015     &    23.831       &    0.738      &     2.081       &       31.524          & 0.738     \\
NRDF~\cite{he2024nrdf}         & 1.306        & 10.388      & 0.132 &     0.899     &    6.705        &    0.125      &      0.964      &  10.388  & 0.137\\
DPoser~\cite{lu2023dposer}    & 0.172        & 15.250               &   0.082    &     0.185     &    15.865       &    0.082      &       0.214           &   14.691       &   0.087   \\
\method{} (Ours)              & 0.171 & 15.456      & 0.079 &     0.237     &    13.197       &    0.078      &   0.161               &    15.828      &  0.090   \\ \hline
\end{tabular}
        \caption{Results of fitting to partial joint observations and analyzing FID, APD, and $d_\text{NN}$ results.}
        \label{table:ik}
\end{table*}
\subsection{Inverse Kinematics for Partial Observations}
\begin{figure}[!h]
    \centering
    \includegraphics[width=.9\linewidth]{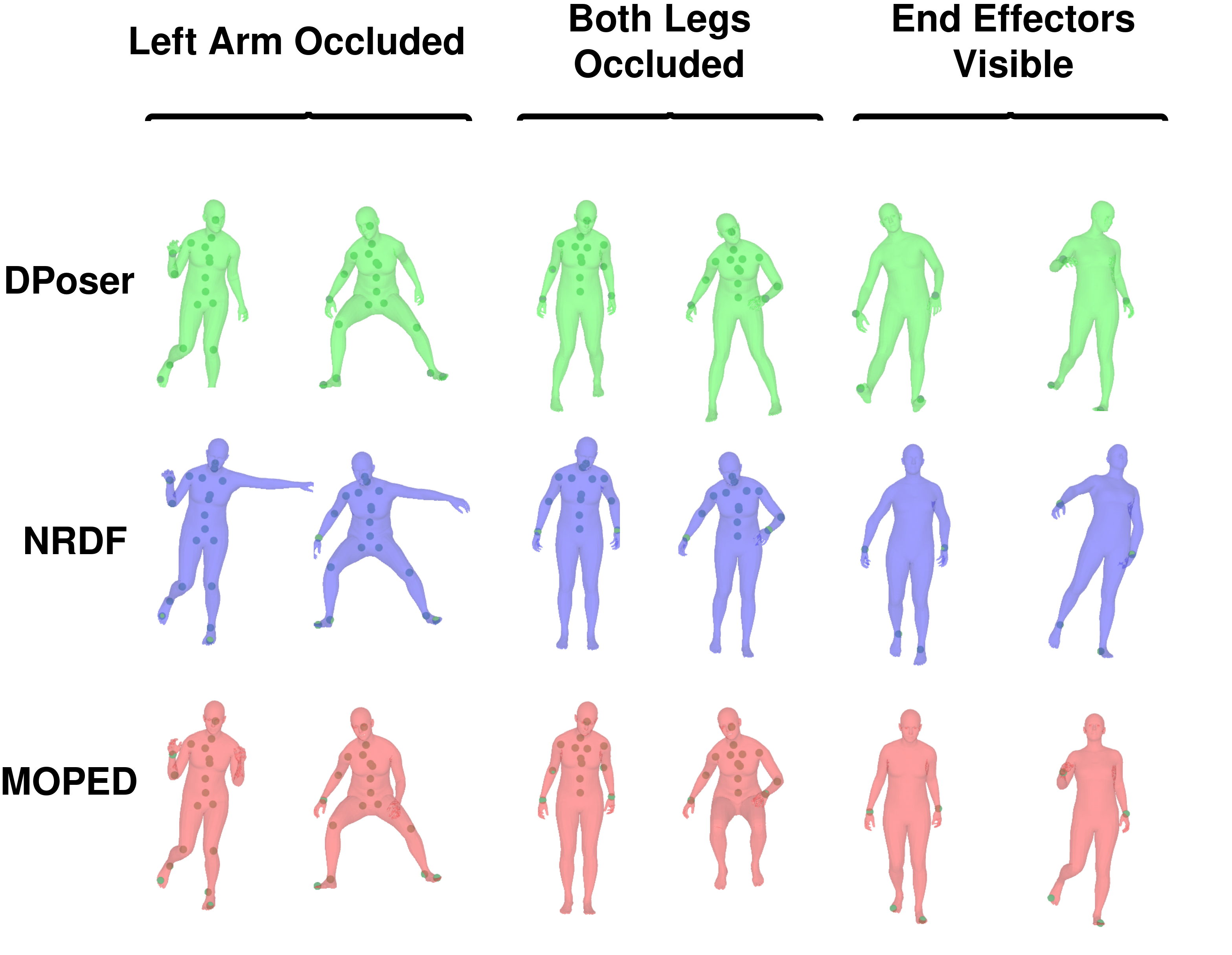}
    \caption{\method{} is capable of more realistic and diverse pose completion from sparse observations. We observe that when the arm is occluded we find that NRDF tends to produce the same results. Meanwhile, when only the end effectors are visible DPoser tends to produce poses that visually look unnatural. \method{} however is able to handle all three cases and produce realistic poses. Additional results with larger images and text-conditioned pose completion are in the Supplemental Material.}
    \label{fig:PoseCompletion}
\end{figure}
One important use of pose priors is the ability to refine pose acquisition from sensors. This is a relatively expensive process and cost-saving measures such as sparse sensor placement or partial observations are under-constrained problems. This is where pose priors can help improve inverse kinematics solvers for more accurate poses. For this experiment, we sample the AMASS test split and compute the 3D joints from the ground truth SMPL parameters. We than apply a mask based off the simulated occlusion we are testing. Specifically we test the following scenarios, legs occluded, single arm occluded, and whole body occlusion where only the end effectors are visible. We use \method{} as a prior similar to the SMPLify framework. We optimize the data term $\mathcal{L}_\text{Data}$ but instead of using the 2D key-points we are using the observed 3D sparse joints which is defined as:
\begin{align}
    \mathcal{L}_\text{Data} = \sum_\text{joint i} ||R_\theta(J(\beta))_i - J_\text{obv,i}||_2^2
    \label{eqn:data}
\end{align}
In this setting we set our DDIM schedule from $t=900$ to $t=10$ and use a linearly spaced schedule over $400$ steps. We report quantitative results in Table.~\ref{table:ik} using FID, APD, and $d_\text{NN}$ metrics and compare against PoseNDF, NRDF, and DPoser. From our experiments we observed that \method{} and DPoser are able to achieve far better FID and $d_\text{NN}$ scores than NRDF which is the current state of the art pose prior. Additionally both diffusion based methods are able to maintain far more diversity in terms of APD over NRDF. In the occluded arm and the visible end effectors experiment we are able to achieve similar if not better realism when compared to DPoser in terms of FID and $d_\text{NN}$ and maintaining similar if not higher diversity in terms of APD. Visually, in Fig. \ref{fig:PoseCompletion} we can see that \method{} is consistently generating more diverse and realistic poses when compared to previous works such as NRDF and DPoser. Despite DPoser having slightly better numerical results for the occluded legs case, we find that visually DPoser tends to produce variations of a standing pose while MOPED is able to produce more varied and realistic leg positions. 

\subsection{Ablation Studies}
    \begin{table}[!h]
        \centering
        \setlength{\tabcolsep}{1mm}
        \begin{tabular}{lcc}
        \toprule
         & EHF  &3DPW \\
         \cmidrule{2-3}
        \textbf{Method}   & \textbf{PA-MPJPE} & \textbf{PA-MPJPE} \\
        \midrule

        \method{} Unconditional     & 49.40   & 51.1 \\
        \method{} Image    & 49.22   & 50.9 \\
        \method{} Text    & 49.55    & 51.01 \\
        \method{} Joint Image-Text & \textbf{49.05}    & \textbf{50.78} \\
        \bottomrule
        \end{tabular}        
        \caption{We examine the performance of \method{} for pose estimation in an unconditional, image conditioned, text conditioned, and joint image and text conditioned settings.}
        \label{table:hmr_abl}
    \end{table}

\noindent\textbf{Human Mesh Regression:} In order to demonstrate the effectiveness of the inclusion of conditioning information, we test \method{} under different conditioning inputs. From Table \ref{table:hmr_abl}, we can see that \method{} in an unconditional setting offers competitive performance when compared to previous works, while the inclusion of image features results in improvements across the board. The inclusion of text features can also improve results as seen in the 3DPW setting. The decrease in performance in EHF datset can likely be attributed to the quality of the captions as all text captions in test settings are automatically generated by an off the shelf image captioning model \cite{li2022blip}. Interestingly, the inclusion of both features however results in a small synergistic improvement for both datasets regardless of the performance in the text only setting. This may be due to the fact that HMR2.0 predictions are already fairly accurate and as such any additional contextual information will result in minor improvements. We observe that with a less accurate initialization in the SMPLify framework, conditioning information can provide a much larger improvement with additional experiments in the Supplemental Material. While the quantitative improvements are relatively small, the main advantage of \method{} is the flexibility for a wide range of tasks and its ability to work with or without any conditioning. Also, as shown in Fig. \ref{fig:Generation Examples}, conditioning leads to much more realistic poses than the unconditional case. 

\section{Conclusion}
\label{sec:con}
We propose \method, a novel multi-modal transformer based diffusion pose prior with a flexible conditioning mechanism and wide applicability for a number of downstream tasks. Through extensive quantitative and qualitative experiments, we demonstrate that \method{} learns an improved distribution of human poses when compared to existing human pose prior models. We show that when \method{} is used as a prior for 3D human mesh regression, it outperforms existing pose priors and can refine state of the art pose estimation models by allowing the incorporation of visual/and or textual features into the refinement process. In the task of pose generation, \method{} offers superior performance over existing works demonstrating it's superior realism and diversity. Finally, we demonstrate \method{}'s superior performance for pose completion from partial observations.


\bibliography{main}

\begin{thebibliography}{48}
\providecommand{\natexlab}[1]{#1}

\bibitem[{Aliakbarian et~al.(2020)Aliakbarian, Saleh, Salzmann, Petersson, and Gould}]{aliakbarian2020stochastic}
Aliakbarian, S.; Saleh, F.~S.; Salzmann, M.; Petersson, L.; and Gould, S. 2020.
\newblock A stochastic conditioning scheme for diverse human motion prediction.
\newblock In \emph{Proceedings of the IEEE/CVF Conference on Computer Vision and Pattern Recognition}, 5223--5232.

\bibitem[{Andriluka et~al.(2014)Andriluka, Pishchulin, Gehler, and Schiele}]{andriluka14cvpr}
Andriluka, M.; Pishchulin, L.; Gehler, P.; and Schiele, B. 2014.
\newblock 2D Human Pose Estimation: New Benchmark and State of the Art Analysis.
\newblock In \emph{IEEE Conference on Computer Vision and Pattern Recognition (CVPR)}.

\bibitem[{Andriluka, Roth, and Schiele(2009)}]{andriluka2009pictorial}
Andriluka, M.; Roth, S.; and Schiele, B. 2009.
\newblock Pictorial structures revisited: People detection and articulated pose estimation.
\newblock In \emph{2009 IEEE conference on computer vision and pattern recognition}, 1014--1021. IEEE.

\bibitem[{Bogo et~al.(2016)Bogo, Kanazawa, Lassner, Gehler, Romero, and Black}]{bogo2016keep}
Bogo, F.; Kanazawa, A.; Lassner, C.; Gehler, P.; Romero, J.; and Black, M.~J. 2016.
\newblock Keep it SMPL: Automatic estimation of 3D human pose and shape from a single image.
\newblock In \emph{Computer Vision--ECCV 2016: 14th European Conference, Amsterdam, The Netherlands, October 11-14, 2016, Proceedings, Part V 14}, 561--578. Springer.

\bibitem[{{Cao} et~al.(2019){Cao}, {Hidalgo Martinez}, {Simon}, {Wei}, and {Sheikh}}]{8765346}
{Cao}, Z.; {Hidalgo Martinez}, G.; {Simon}, T.; {Wei}, S.; and {Sheikh}, Y.~A. 2019.
\newblock OpenPose: Realtime Multi-Person 2D Pose Estimation using Part Affinity Fields.
\newblock \emph{IEEE Transactions on Pattern Analysis and Machine Intelligence}.

\bibitem[{Chung et~al.(2022)Chung, Kim, Mccann, Klasky, and Ye}]{chung2022diffusion}
Chung, H.; Kim, J.; Mccann, M.~T.; Klasky, M.~L.; and Ye, J.~C. 2022.
\newblock Diffusion posterior sampling for general noisy inverse problems.
\newblock \emph{arXiv preprint arXiv:2209.14687}.

\bibitem[{Ci et~al.(2023)Ci, Wu, Zhu, Ma, Dong, Zhong, and Wang}]{10204476}
Ci, H.; Wu, M.; Zhu, W.; Ma, X.; Dong, H.; Zhong, F.; and Wang, Y. 2023.
\newblock GFPose: Learning 3D Human Pose Prior with Gradient Fields.
\newblock In \emph{2023 IEEE/CVF Conference on Computer Vision and Pattern Recognition (CVPR)}, 4800--4810. Los Alamitos, CA, USA: IEEE Computer Society.

\bibitem[{Davydov et~al.(2022)Davydov, Remizova, Constantin, Honari, Salzmann, and Fua}]{davydov2022adversarial}
Davydov, A.; Remizova, A.; Constantin, V.; Honari, S.; Salzmann, M.; and Fua, P. 2022.
\newblock Adversarial parametric pose prior.
\newblock In \emph{Proceedings of the IEEE/CVF Conference on Computer Vision and Pattern Recognition}, 10997--11005.

\bibitem[{Dutta et~al.(2024)Dutta, Lal, Raychaudhuri, Ta, and Roy-Chowdhury}]{dutta2024poise}
Dutta, A.; Lal, R.; Raychaudhuri, D.~S.; Ta, C.-K.; and Roy-Chowdhury, A.~K. 2024.
\newblock POISE: Pose Guided Human Silhouette Extraction under Occlusions.
\newblock In \emph{Proceedings of the IEEE/CVF Winter Conference on Applications of Computer Vision}, 6153--6163.

\bibitem[{Felzenszwalb and Huttenlocher(2005)}]{felzenszwalb2005pictorial}
Felzenszwalb, P.~F.; and Huttenlocher, D.~P. 2005.
\newblock Pictorial structures for object recognition.
\newblock \emph{International journal of computer vision}, 61: 55--79.

\bibitem[{Geman(1987)}]{geman1987statistical}
Geman, S. 1987.
\newblock Statistical methods for tomographic image restoration.
\newblock \emph{Bull. Internat. Statist. Inst.}, 52: 5--21.

\bibitem[{Goel et~al.(2023)Goel, Pavlakos, Rajasegaran, Kanazawa*, and Malik*}]{goel2023humans}
Goel, S.; Pavlakos, G.; Rajasegaran, J.; Kanazawa*, A.; and Malik*, J. 2023.
\newblock Humans in 4{D}: Reconstructing and Tracking Humans with Transformers.
\newblock In \emph{International Conference on Computer Vision (ICCV)}.

\bibitem[{Gong et~al.(2023)Gong, Foo, Fan, Ke, Rahmani, and Liu}]{gong2023diffpose}
Gong, J.; Foo, L.~G.; Fan, Z.; Ke, Q.; Rahmani, H.; and Liu, J. 2023.
\newblock Diffpose: Toward more reliable 3d pose estimation.
\newblock In \emph{Proceedings of the IEEE/CVF Conference on Computer Vision and Pattern Recognition}, 13041--13051.

\bibitem[{He et~al.(2024)He, Tiwari, Birdal, Lenssen, and Pons-Moll}]{he2024nrdf}
He, Y.; Tiwari, G.; Birdal, T.; Lenssen, J.~E.; and Pons-Moll, G. 2024.
\newblock Nrdf: Neural riemannian distance fields for learning articulated pose priors.
\newblock In \emph{Proceedings of the IEEE/CVF Conference on Computer Vision and Pattern Recognition}, 1661--1671.

\bibitem[{Ho, Jain, and Abbeel(2020)}]{ho2020denoising}
Ho, J.; Jain, A.; and Abbeel, P. 2020.
\newblock Denoising diffusion probabilistic models.
\newblock \emph{Advances in neural information processing systems}, 33: 6840--6851.

\bibitem[{Ho and Salimans(2021)}]{ho2021classifier}
Ho, J.; and Salimans, T. 2021.
\newblock Classifier-Free Diffusion Guidance.
\newblock In \emph{NeurIPS 2021 Workshop on Deep Generative Models and Downstream Applications}.

\bibitem[{Ionescu et~al.(2014)Ionescu, Papava, Olaru, and Sminchisescu}]{h36m_pami}
Ionescu, C.; Papava, D.; Olaru, V.; and Sminchisescu, C. 2014.
\newblock Human3.6M: Large Scale Datasets and Predictive Methods for 3D Human Sensing in Natural Environments.
\newblock \emph{IEEE Transactions on Pattern Analysis and Machine Intelligence}, 36(7): 1325--1339.

\bibitem[{Kanazawa et~al.(2018)Kanazawa, Black, Jacobs, and Malik}]{hmrKanazawa17}
Kanazawa, A.; Black, M.~J.; Jacobs, D.~W.; and Malik, J. 2018.
\newblock End-to-end Recovery of Human Shape and Pose.
\newblock In \emph{Computer Vision and Pattern Recognition (CVPR)}.

\bibitem[{Li et~al.(2023)Li, Shi, Dai, Zheng, Wang, Sun, Guo, Li, Zou, and Xiong}]{li2023pose}
Li, H.; Shi, B.; Dai, W.; Zheng, H.; Wang, B.; Sun, Y.; Guo, M.; Li, C.; Zou, J.; and Xiong, H. 2023.
\newblock Pose-oriented transformer with uncertainty-guided refinement for 2d-to-3d human pose estimation.
\newblock In \emph{Proceedings of the AAAI Conference on Artificial Intelligence}, volume~37, 1296--1304.

\bibitem[{Li et~al.(2022{\natexlab{a}})Li, Li, Xiong, and Hoi}]{li2022blip}
Li, J.; Li, D.; Xiong, C.; and Hoi, S. 2022{\natexlab{a}}.
\newblock BLIP: Bootstrapping Language-Image Pre-training for Unified Vision-Language Understanding and Generation.
\newblock In \emph{ICML}.

\bibitem[{Li et~al.(2022{\natexlab{b}})Li, Liu, Zhang, Xu, and Yan}]{li2022cliff}
Li, Z.; Liu, J.; Zhang, Z.; Xu, S.; and Yan, Y. 2022{\natexlab{b}}.
\newblock CLIFF: Carrying Location Information in Full Frames into Human Pose and Shape Estimation.
\newblock In \emph{ECCV}.

\bibitem[{Lin, Wang, and Liu(2021)}]{lin2021end-to-end}
Lin, K.; Wang, L.; and Liu, Z. 2021.
\newblock End-to-End Human Pose and Mesh Reconstruction with Transformers.
\newblock In \emph{CVPR}.

\bibitem[{Lin et~al.(2014)Lin, Maire, Belongie, Hays, Perona, Ramanan, Doll{\'a}r, and Zitnick}]{lin2014microsoft}
Lin, T.-Y.; Maire, M.; Belongie, S.; Hays, J.; Perona, P.; Ramanan, D.; Doll{\'a}r, P.; and Zitnick, C.~L. 2014.
\newblock Microsoft coco: Common objects in context.
\newblock In \emph{Computer Vision--ECCV 2014: 13th European Conference, Zurich, Switzerland, September 6-12, 2014, Proceedings, Part V 13}, 740--755. Springer.

\bibitem[{Loper et~al.(2015)Loper, Mahmood, Romero, Pons-Moll, and Black}]{10.1145/2816795.2818013}
Loper, M.; Mahmood, N.; Romero, J.; Pons-Moll, G.; and Black, M.~J. 2015.
\newblock SMPL: a skinned multi-person linear model.
\newblock \emph{ACM Trans. Graph.}, 34(6).

\bibitem[{Lu et~al.(2023)Lu, Lin, Dou, Zhang, Deng, and Wang}]{lu2023dposer}
Lu, J.; Lin, J.; Dou, H.; Zhang, Y.; Deng, Y.; and Wang, H. 2023.
\newblock DPoser: Diffusion Model as Robust 3D Human Pose Prior.
\newblock \emph{arXiv preprint arXiv:2312.05541}.

\bibitem[{Ma et~al.(2017)Ma, Jia, Sun, Schiele, Tuytelaars, and Van~Gool}]{ma2017pose}
Ma, L.; Jia, X.; Sun, Q.; Schiele, B.; Tuytelaars, T.; and Van~Gool, L. 2017.
\newblock Pose guided person image generation.
\newblock \emph{Advances in neural information processing systems}, 30.

\bibitem[{Mahmood et~al.(2019)Mahmood, Ghorbani, Troje, Pons-Moll, and Black}]{AMASS:ICCV:2019}
Mahmood, N.; Ghorbani, N.; Troje, N.~F.; Pons-Moll, G.; and Black, M.~J. 2019.
\newblock {AMASS}: Archive of Motion Capture as Surface Shapes.
\newblock In \emph{International Conference on Computer Vision}, 5442--5451.

\bibitem[{Marchand, Uchiyama, and Spindler(2016)}]{Marchand2016PoseEF}
Marchand, {\'E}.; Uchiyama, H.; and Spindler, F. 2016.
\newblock Pose Estimation for Augmented Reality: A Hands-On Survey.
\newblock \emph{IEEE Transactions on Visualization and Computer Graphics}, 22: 2633--2651.

\bibitem[{Martinez et~al.(2017)Martinez, Hossain, Romero, and Little}]{martinez2017simple}
Martinez, J.; Hossain, R.; Romero, J.; and Little, J.~J. 2017.
\newblock A simple yet effective baseline for 3d human pose estimation.
\newblock In \emph{Proceedings of the IEEE international conference on computer vision}, 2640--2649.

\bibitem[{Mehta et~al.(2017)Mehta, Rhodin, Casas, Fua, Sotnychenko, Xu, and Theobalt}]{mono-3dhp2017}
Mehta, D.; Rhodin, H.; Casas, D.; Fua, P.; Sotnychenko, O.; Xu, W.; and Theobalt, C. 2017.
\newblock Monocular 3D Human Pose Estimation In The Wild Using Improved CNN Supervision.
\newblock In \emph{3D Vision (3DV), 2017 Fifth International Conference on}. IEEE.

\bibitem[{Moon and Lee(2020)}]{Moon_2020_ECCV_I2L-MeshNet}
Moon, G.; and Lee, K.~M. 2020.
\newblock I2L-MeshNet: Image-to-Lixel Prediction Network for Accurate 3D Human Pose and Mesh Estimation from a Single RGB Image.
\newblock In \emph{European Conference on Computer Vision (ECCV)}.

\bibitem[{Parameswaran and Chellappa(2004)}]{parameswaran2004view}
Parameswaran, V.; and Chellappa, R. 2004.
\newblock View independent human body pose estimation from a single perspective image.
\newblock In \emph{Proceedings of the 2004 IEEE Computer Society Conference on Computer Vision and Pattern Recognition, 2004. CVPR 2004.}, volume~2, II--II. IEEE.

\bibitem[{Pavlakos et~al.(2019)Pavlakos, Choutas, Ghorbani, Bolkart, Osman, Tzionas, and Black}]{SMPL-X:2019}
Pavlakos, G.; Choutas, V.; Ghorbani, N.; Bolkart, T.; Osman, A. A.~A.; Tzionas, D.; and Black, M.~J. 2019.
\newblock Expressive Body Capture: 3D Hands, Face, and Body from a Single Image.
\newblock In \emph{Proceedings IEEE Conf. on Computer Vision and Pattern Recognition (CVPR)}.

\bibitem[{Punnakkal et~al.(2021)Punnakkal, Chandrasekaran, Athanasiou, Quiros-Ramirez, and Black}]{BABEL:CVPR:2021}
Punnakkal, A.~R.; Chandrasekaran, A.; Athanasiou, N.; Quiros-Ramirez, A.; and Black, M.~J. 2021.
\newblock {BABEL}: Bodies, Action and Behavior with English Labels.
\newblock In \emph{Proceedings IEEE/CVF Conf.~on Computer Vision and Pattern Recognition (CVPR)}, 722--731.

\bibitem[{Radford et~al.(2021)Radford, Kim, Hallacy, Ramesh, Goh, Agarwal, Sastry, Askell, Mishkin, Clark et~al.}]{radford2021learning}
Radford, A.; Kim, J.~W.; Hallacy, C.; Ramesh, A.; Goh, G.; Agarwal, S.; Sastry, G.; Askell, A.; Mishkin, P.; Clark, J.; et~al. 2021.
\newblock Learning transferable visual models from natural language supervision.
\newblock In \emph{International conference on machine learning}, 8748--8763. PMLR.

\bibitem[{Raychaudhuri et~al.(2023)Raychaudhuri, Ta, Dutta, Lal, and Roy-Chowdhury}]{raychaudhuri2023prior}
Raychaudhuri, D.~S.; Ta, C.-K.; Dutta, A.; Lal, R.; and Roy-Chowdhury, A.~K. 2023.
\newblock Prior-guided Source-free Domain Adaptation for Human Pose Estimation.
\newblock In \emph{Proceedings of the IEEE/CVF International Conference on Computer Vision}, 14996--15006.

\bibitem[{Rombach et~al.(2021)Rombach, Blattmann, Lorenz, Esser, and Ommer}]{rombach2021high}
Rombach, R.; Blattmann, A.; Lorenz, D.; Esser, P.; and Ommer, B. 2021.
\newblock High-resolution image synthesis with latent diffusion models. 2022 IEEE.
\newblock In \emph{CVF Conference on Computer Vision and Pattern Recognition (CVPR)}, 10674--10685.

\bibitem[{Sohl-Dickstein et~al.(2015)Sohl-Dickstein, Weiss, Maheswaranathan, and Ganguli}]{sohl2015deep}
Sohl-Dickstein, J.; Weiss, E.; Maheswaranathan, N.; and Ganguli, S. 2015.
\newblock Deep unsupervised learning using nonequilibrium thermodynamics.
\newblock In \emph{International conference on machine learning}, 2256--2265. PMLR.

\bibitem[{Song, Meng, and Ermon(2020)}]{song2020denoising}
Song, J.; Meng, C.; and Ermon, S. 2020.
\newblock Denoising diffusion implicit models.
\newblock \emph{arXiv preprint arXiv:2010.02502}.

\bibitem[{Song et~al.(2021)Song, Sohl-Dickstein, Kingma, Kumar, Ermon, and Poole}]{song2021scorebased}
Song, Y.; Sohl-Dickstein, J.; Kingma, D.~P.; Kumar, A.; Ermon, S.; and Poole, B. 2021.
\newblock Score-Based Generative Modeling through Stochastic Differential Equations.
\newblock In \emph{International Conference on Learning Representations}.

\bibitem[{Stathopoulos, Han, and Metaxas(2024)}]{stathopoulos2024score}
Stathopoulos, A.; Han, L.; and Metaxas, D. 2024.
\newblock Score-Guided Diffusion for 3D Human Recovery.
\newblock In \emph{Proceedings of the IEEE/CVF Conference on Computer Vision and Pattern Recognition}, 906--915.

\bibitem[{Tiwari et~al.(2022)Tiwari, Anti{\'c}, Lenssen, Sarafianos, Tung, and Pons-Moll}]{tiwari2022pose}
Tiwari, G.; Anti{\'c}, D.; Lenssen, J.~E.; Sarafianos, N.; Tung, T.; and Pons-Moll, G. 2022.
\newblock Pose-ndf: Modeling human pose manifolds with neural distance fields.
\newblock In \emph{European Conference on Computer Vision}, 572--589. Springer.

\bibitem[{Vaswani et~al.(2017)Vaswani, Shazeer, Parmar, Uszkoreit, Jones, Gomez, Kaiser, and Polosukhin}]{vaswani2017attention}
Vaswani, A.; Shazeer, N.; Parmar, N.; Uszkoreit, J.; Jones, L.; Gomez, A.~N.; Kaiser, {\L}.; and Polosukhin, I. 2017.
\newblock Attention is all you need.
\newblock \emph{Advances in neural information processing systems}, 30.

\bibitem[{von Marcard et~al.(2018)von Marcard, Henschel, Black, Rosenhahn, and Pons-Moll}]{vonMarcard2018}
von Marcard, T.; Henschel, R.; Black, M.; Rosenhahn, B.; and Pons-Moll, G. 2018.
\newblock Recovering Accurate 3D Human Pose in The Wild Using IMUs and a Moving Camera.
\newblock In \emph{European Conference on Computer Vision (ECCV)}.

\bibitem[{Zhang et~al.(2021)Zhang, Tian, Zhou, Ouyang, Liu, Wang, and Sun}]{pymaf2021}
Zhang, H.; Tian, Y.; Zhou, X.; Ouyang, W.; Liu, Y.; Wang, L.; and Sun, Z. 2021.
\newblock PyMAF: 3D Human Pose and Shape Regression with Pyramidal Mesh Alignment Feedback Loop.
\newblock In \emph{Proceedings of the IEEE International Conference on Computer Vision}.

\bibitem[{Zhang et~al.(2019)Zhang, Wong, Kankanhalli, and Geng}]{zhang2019unsupervised}
Zhang, X.; Wong, Y.; Kankanhalli, M.~S.; and Geng, W. 2019.
\newblock Unsupervised domain adaptation for 3D human pose estimation.
\newblock In \emph{Proceedings of the 27th ACM International Conference on Multimedia}, 926--934.

\bibitem[{Zhou et~al.(2019)Zhou, Barnes, Jingwan, Jimei, and Hao}]{Zhou_2019_CVPR}
Zhou, Y.; Barnes, C.; Jingwan, L.; Jimei, Y.; and Hao, L. 2019.
\newblock On the Continuity of Rotation Representations in Neural Networks.
\newblock In \emph{The IEEE Conference on Computer Vision and Pattern Recognition (CVPR)}.

\bibitem[{Zhu, Zheng, and Nevatia(2023)}]{zhu2023gait}
Zhu, H.; Zheng, Z.; and Nevatia, R. 2023.
\newblock Gait recognition using 3-d human body shape inference.
\newblock In \emph{Proceedings of the IEEE/CVF Winter Conference on Applications of Computer Vision}, 909--918.

\end{thebibliography}
\clearpage
\end{document}